\documentclass[10pt,twocolumn,letterpaper]{article}

\usepackage{iccv}
\usepackage{times}
\usepackage{epsfig}
\usepackage{graphicx}
\usepackage{amsmath}
\usepackage{amssymb}
\usepackage[dvipsnames]{xcolor}
\usepackage{eso-pic}
\usepackage{booktabs}
\usepackage{xcolor}
\usepackage{color, colortbl}
\usepackage{algorithm}
\usepackage{algorithmicx}
\usepackage{algpseudocode}
\definecolor{Gray}{gray}{0.9}

\usepackage[breaklinks=true,colorlinks,bookmarks=false]{hyperref}

\iccvfinalcopy 

\def\iccvPaperID{****} 

\begin{document}

\title{FoSp: Focus and Separation Network for Early Smoke Segmentation 
}

\author{Lujian Yao, Haitao Zhao, Jingchao Peng, Zhongze Wang, Kaijie Zhao\\
Automation Department, School of Information Science and Engineering,\\
East China University of Science and Technology\\
}

\maketitle

\begin{abstract}

Early smoke segmentation (ESS) enables the accurate identification of smoke sources, facilitating the prompt extinguishing of fires and preventing large-scale gas leaks. But ESS poses greater challenges than conventional object and regular smoke segmentation due to its small scale and transparent appearance, which can result in high miss detection rate and low precision. To address these issues, a Focus and Separation Network (FoSp) is proposed. We first introduce a \textit{Focus} module employing bidirectional cascade which guides low-resolution and high-resolution features towards mid-resolution to locate and determine the scope of smoke, reducing the \textit{miss detection rate}. Next, we propose a \textit{Separation} module that separates smoke images into a pure smoke foreground and a smoke-free background, enhancing the contrast between smoke and background fundamentally, improving segmentation \textit{precision}. Finally, a \textit{Domain Fusion} module is developed to integrate the distinctive features of the two modules which can balance \textit{recall} and \textit{precision} to achieve high $F_{\beta}$. 
Futhermore, to promote the development of ESS, we introduce a high-quality real-world dataset called SmokeSeg, which contains more small and transparent smoke images than the existing datasets. Experimental results show that our model achieves the best performance on three available datasets: SYN70K ($mIoU$: 83.00\%), SMOKE5K ($F_{\beta}$: 81.6\%) and SmokeSeg ($F_{\beta}$: 72.05\%). 
Especially, our FoSp outperforms SegFormer by 7.71\% ($F_{\beta}$) for early smoke segmentation on SmokeSeg.

\end{abstract}

\section{Introduction}

    In wildlife, smoke is an important indicator of fire. Early smoke segmentation (ESS) enables rapid identification of the location of the fire~\cite{muhammad2018early, robinson1979smoke}, facilitating the timely extinguishing of the flames by rescue personnel and preventing the occurrence of large fires. In industrial production, ESS can also aid in promptly detecting the location of gas leaks and prevent the spread of toxic and harmful gases~\cite{hsu2021project}.

    \begin{figure}[!ht]
    \begin{center}
        \includegraphics[width=\linewidth]{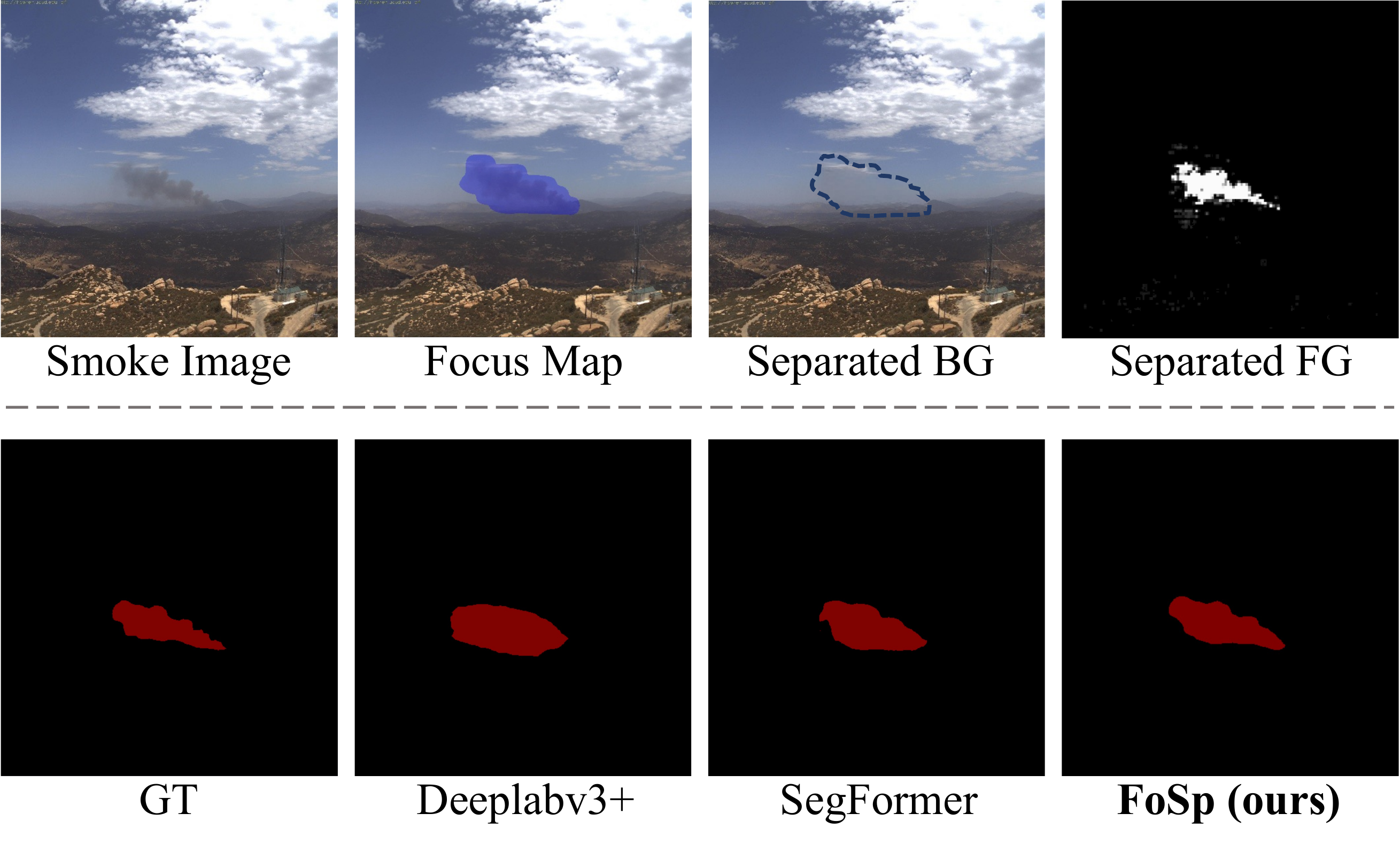}
    \end{center}
    \vspace{-3mm}
       \caption{\textbf{Process and Comparison.} A Focus Map is generated by the proposed \textit{Focus} module to determine the smoke scope. Then, we introduce a \textit{Separation} module to extract the \textit{smoke foreground} (Separated FG) from the \textit{smoke-free background} (Separated BG), thus enhancing the contrast between the foreground and the background. This method leads to obtaining a more \textit{refined} smoke map than strong baselines like DeepLabV3+\cite{chen2018encoder} and SegFormer\cite{xie2021segformer}.}
    \label{fig:first}
    \vspace{-5mm}
    \end{figure}
    Smoke segmentation presents a more challenging task than ordinary object segmentation. Unlike conventional objects, which are typically rigid and have clear edges, smoke is a non-rigid substance with ever-changing form and blurred edges. 
    Furthermore, early smoke segmentation is much more challenging than regular smoke segmentation, as early smoke is small and transparent, resulting in \textit{high miss detection rate} (incomplete segmentation) and \textit{low precision} (rough-edge segmentation). 
    Previous regular smoke segmentation methods~\cite{ yuan2022cubic, yuan2019wave, yuan2019deep} primarily emphasize larger receptive fields to cope with the variability and blurred edges of smoke. 
    However, these methods do not work well for identifying early smoke with small scale and transparent appearance. 

    To address the issue of early smoke, we propose a \textit{Focus and Separation Network} (FoSp) which concentrates on the formulation of smoke images.
    Firstly, we design a \textit{Focus} module utilizing bidirctional cascade which directs low-resolution and high-resolution features towards mid-resolution
    to locate and confirm the scope of smoke, effectively reducing the \textit{miss detection rate}. Subsequently, 
    we propose a \textit{Separation} module that separates smoke images into a pure smoke foreground and a smoke-free background, increasing the contrast between smoke and background to improve the \textit{precision} of segmentation. 
    Finally, we integrate the distinctive features of the two modules and develop a \textit{Domain Fusion} module that balances \textit{recall} and \textit{precision} to obtain the general $F_{\beta}$.
    
    Specifically, 1) \textit{Focus module}: We propose a Bidirectional Cascade Generator (BCG) for focus module. 
    BCG employs bidirectional guidance of low-resolution and high-resolution features, targeting respectively the smoke body and the smoke edge, towards mid-resolution features to generate a Focus Map (FM) that can fully encompass the smoke region.
    2) \textit{Separation module}: In order to retain information and maintain consistency with high-level segmentation tasks, we adopt a feature-level approach to separate the foreground and background of smoke images. 
    By utilizing FM, we acquire the smoke region and adopt the  inpainting technique to complete the smoke area with regular smoke-free background, using the adjacent environment (especially the area surrounding the smoke) as a reference. We subsequently deduct the smoke-free background from the initial smoke region to obtain the separated smoke foreground. 3) \textit{Domain Fusion}: 
    Since the foreground features are generated by the inpainter (inpainting model), the origin features and foreground features actually belong to distinct feature domains.
    An MLP at each stage is designed to hierarchically merge the features from two domains.
    
    Furthermore, we provide a large and real-world dataset called SmokeSeg, which includes 6,144 real images of smoke, with a considerable number of them featuring small and transparent smoke. SmokeSeg currently boasts the largest number of real images among publicly available smoke segmentation datasets. Notably, the SmokeSeg contains 4.5 times more real images than SMOKE5K~\cite{yan2022transmission}, which only comprises 1,360 real images. 
    
    Our main contributions can be summarized as follows:

\begin{itemize}
    \item We propose a Focus and Separation Network (FoSp), where the Focus module and Separation module are introduced to reduce the miss detection rate and improve precision of early smoke segmentation, respectively.
    \item We have created a large-scale dataset named \textit{SmokeSeg} for early smoke segmentation, which includes 6144 real images with pixel-wise annotations.
    \item Our FoSp achieve the best performance on three datasets: SYN70K ($mIoU$: 83.00\%), SMOKE5K ($F_{\beta}$: 81.6\%) and SmokeSeg ($F_{\beta}$: 72.05\%). Particularly, FoSp surpasses SegFormer by 7.71\% ($F_{\beta}$) for early smoke segmentation on SmokeSeg.
\end{itemize}

\section{Related Work}
    In this section, we begin by reviewing smoke segmentation methods that are based on both traditional and deep learning approaches. Subsequently, we provide a brief overview of prior attention and conclude by summarizing current status of smoke segmentation datasets.

    \textbf{Early Smoke Segmentation:}
    Although current semantic segmentation methods ~\cite{chen2018encoder, strudel2021segmenter, xie2021segformer} are effective at segmenting regular objects (i.e., those with clear outlines and roughly the same shape), these generic algorithms are not suitable for early smoke segmentation due to its varying transparency and small scale.
    Traditional smoke segmentation methods have mainly focused on extracting high-quality color and texture features~\cite{ mahmoud2019forest, xing2015smoke, yuan2019learning}, and deep-learning-based methods~\cite{yuan2022cubic, yuan2019wave, yuan2019deep} mainly focus on extracting features with larger receptive fields to handle the variability and blurred edges of smoke. 
    However, these methods suffer from high miss detection rate and low precision in segmenting small and transparent early smoke. And many of them~\cite{yuan2019wave, yuan2019deep} have only been quantitatively evaluated on synthetic datasets. Despite some methods~\cite{jia2019automatic, wang2014early} claiming to address early smoke, they still rely on images of regular smoke.
    Therefore, the existing smoke segmentation methods are not capable of effectively addressing the issue of early smoke.

    \begin{figure*}
        \begin{center}
            \includegraphics[width=0.90\linewidth]{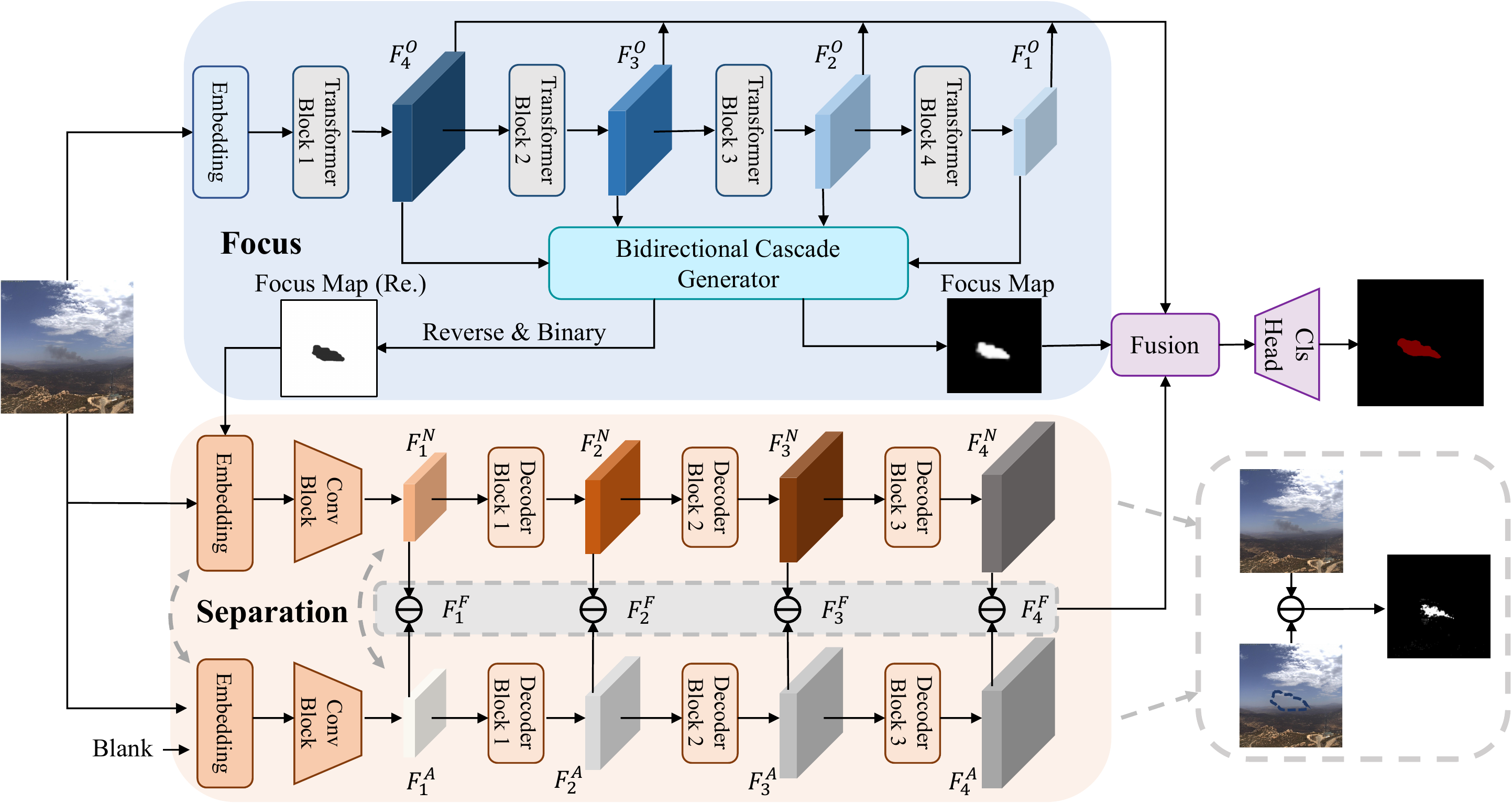}
        \end{center}
        \vspace{-2mm}
        \caption{\textbf{The structure of our FoSp.} The Focus module extracts image features and generates a Focus Map using Bidirectional Cascade Generator (BCG). The Separation module consists of two inpainters: one to complete the smoke areas with smoke-free backgrounds using the original image and Focus Map , and the other using the original image and a blank image. The smoke foreground features are obtained by subtracting the two features, and then the origin features, Focus Map, and foreground features are fused in the Domain Fusion module to obtain the final prediction. }
        \label{fig:FoSp_Pipeline}
        \vspace{-3mm}
    \end{figure*}

    \textbf{Prior Attention:} Prior attention is a mechanism that adds a branch to the network before the backbone~\cite{uzkent2020efficient, wang2020glance, xie2020spatially} or before the final predictions~\cite{najibi2019autofocus, yang2022querydet}, making the network focus on a specific area of the image in early to obtain finer predictions. However, current methods are mostly used for object detection and are designed to improve \textit{precision}. How to obtain a prior attention for segmentation with high \textit{recall} is still to be resolved.

    \textbf{Smoke Segmentation Datasets:} 
    Currently, there are two main smoke segmentation datasets: SYN70K~\cite{yuan2019deep} and SMOKE5K~\cite{yan2022transmission}. SYN70K is a synthetic dataset comprising 70k images, while SMOKE5K contains 1k real images and 4k synthetic images, with the latter being selected from SYN70K. However, neither of these datasets is specifically designed for early smoke segmentation. The SYN70K comprises entirely of synthetic smoke, which is larger in scale and significantly different from real images. The SMOKE5K has only a small number of real smoke images and a low proportion of early small and transparent smoke. Therefore, there is currently no dataset particularly designed for early smoke segmentation.

\section{Method}
    In this section, we first present a general introduction of our FoSp. Then we describe the three modules of FoSp: Focus module, Separation module and Domain Fusion module. Finally, we present the loss function used to train our model.
    
    \subsection{Introduction of Focus and Separation (FoSp)}

\noindent\textbf{Intuition.} We assume that the smoke image $\mathbf{I} \in \mathbb{R}^{H \times W \times 3}$ is composed of a smoke-free background $\mathbf{B}  \in \mathbb{R}^{H \times W \times 3}$ and a smoke foreground $\mathbf{S}  \in \mathbb{R}^{H \times W \times 3}$:
    \begin{small}
    \begin{align}
        \label{eq:smoke_formulation}
            \mathbf{I}=(1-\alpha)\odot \mathbf{B} + \alpha \odot \mathbf{S}, 
    \end{align}
\end{small}%
where $\odot$ indicates the element-wise multiplication and the $\alpha \in \mathbb{R}^{H \times W \times 3}$ is used to control the density of the smoke foreground in RGB channels.
Supposing we can obtain a smoke-free background image $\mathbf{B}$, we are capable of enhancing the smoke foreground $\mathbf{S}$ fundamentally by controlling the value of $\alpha$ to obtain a precise segmentation map.
Following such intuition, we propose a \textit{Focus and Separation Network} (FoSp).  

\noindent\textbf{Overview.} As shown in Fig.~\ref{fig:FoSp_Pipeline} , we first propose a Focus module to locate and confirm the scope of smoke. This enables the network to pay closer attention to the smoke and reduce the miss detection rate. Next, a Separation module is introduced to split the smoke image into pure smoke foreground and smoke-free background, which fundamentally enhances the contrast between the foreground and background, thereby improving precision. Finally, a Domain Fusion module is developed to narrow the domain gap between the features generated by the first two modules, ultimately achieving a balance between recall and precision.

    \subsection{Focus Module}
\begin{figure}[t]
    \begin{center}
        \includegraphics[width=0.88\linewidth]{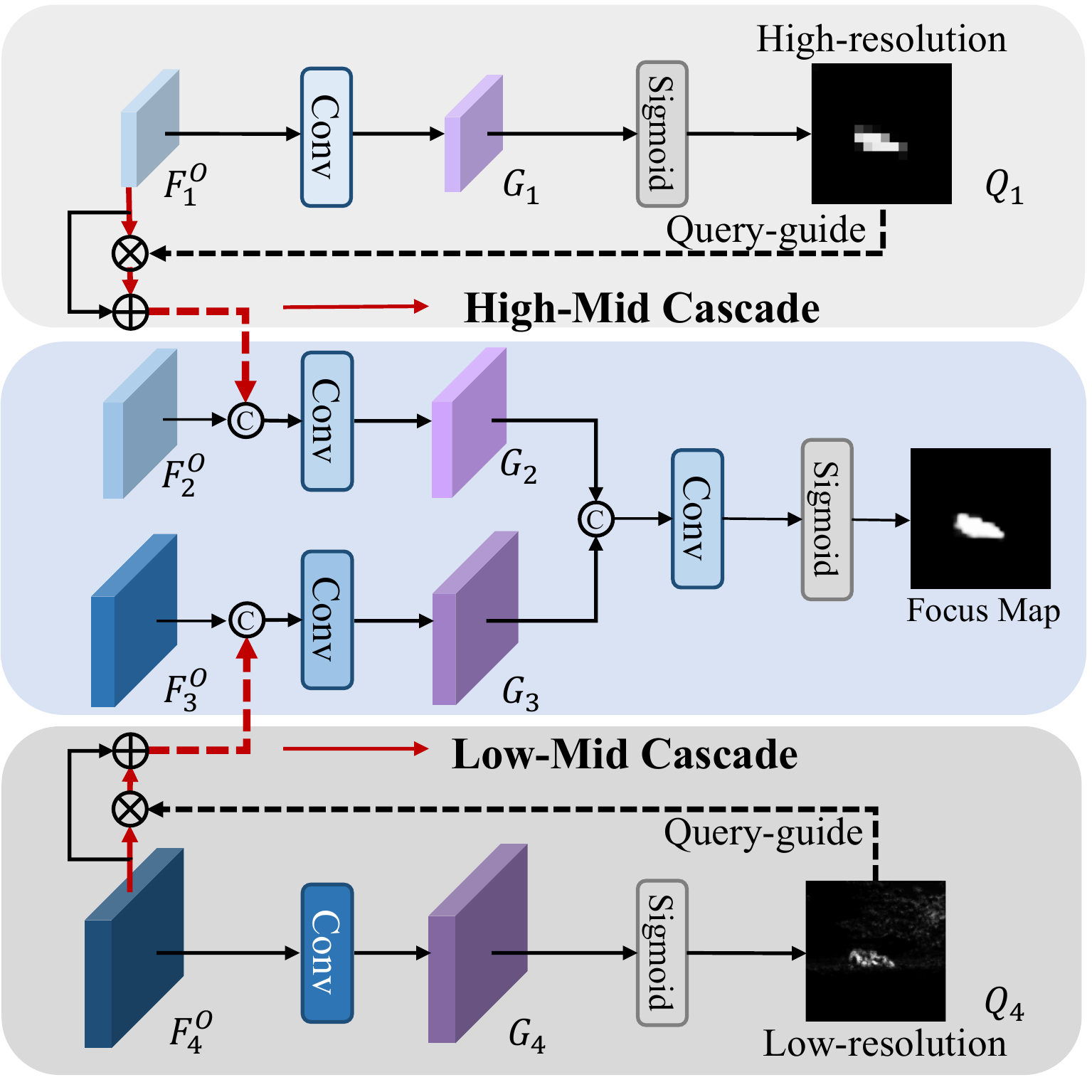}
    \end{center}
    \vspace{-2mm}
       \caption{\textbf{The structure of Bidirectional Cascade Generator.} It guides low-resolution features and high-resolution features towards mid-resolution to obtain a more complete scope of smoke.}
    \label{fig:Focus_Pipeline}
    \vspace{-4mm}
    \end{figure}

    \noindent\textbf{Intuition.} Most segmentation methods use pyramid-based \textit{unidirectional} feature integration to obtain a more refined prediction. However, our Focus module is designed to achieve a more complete smoke area. To achieve this, we introduce a novel \textit{bidirectional} feature cascade approach. We consider the smoke as two parts: the low-opacity smoke body and the high-opacity smoke edge.
    As shown in Fig.~\ref{fig:Focus_Pipeline},
    low-resolution features can provide an approximate outline of smoke body, However, the effectiveness of capturing transparent regions of smoke is compromised. Conversely, high-resolution features offer a more precise attention and can identify transparent parts of smoke edge, but cannot fully capture the entire smoke. Therefore, we propose a bidirectional fusion of features which can integrate features from both low and high resolution, resulting in a complete scope of the smoke. 
    
    \noindent\textbf{Overview.} In the focus module, a Bidirectional Cascade Generator (BCG) is proposed to integrate the original multi-scale image features for locating and confirming the range of smoke, which we refer to as the Focus Map (FM). As show in Fig.~\ref{fig:Focus_Pipeline}, BCG bidirectionally guides low-resolution features and high-resolution features towards mid-resolution to obtain a complete scope of smoke. 
    
\noindent\textbf{Detail.} As shown in Fig~\ref{fig:FoSp_Pipeline}, in the feature extraction section, MiT-B3~\cite{xie2021segformer} is adopted as our backbone, which is a transformer-based backbone with large receptive field that can handle the variability of smoke. Specifically, we preprocess the image using an embedding layer and extract features at four different scales using four Transformer blocks. These scales range from small to large and are respectively $\textbf{F}_{i} \in \mathbb{R}^{\frac{H}{2^{6-i}}\times \frac{W}{2^{6-i}} \times C_{i}}(i=1, 2, 3, 4)$. 
The BCG is illustrated in Fig.~\ref{fig:Focus_Pipeline}, which consists of two parts: Low-Mid Cascade (LMC) and High-Mid Cascade (HMC). The two modules are similar, so we will use the LMC module as an example to explain. In LMC, The lowest-resolution feature $\mathbf{F}_{1}\!\in\!\mathbb{R}^{\frac{H}{32} \times \frac{W}{32} \times 1}$ is fed into a layer of Conv to obtain the logits $\mathbf{G}_{1}\!\in\!\mathbb{R}^{\frac{H}{32} \times \frac{W}{32} \times 1}$ and then applied the $\text{Sigmoid}$ function to obtain the prediction $\mathbf{Q}_{1}\!\in\!\mathbb{R}^{\frac{H}{32} \times \frac{W}{32} \times 1}$ of the $\mathbf{F}_{1}$. 
A query-guide approach is adopted to enhance the feature maps by taking the dot product of $\mathbf{Q}_{1}$ and $\mathbf{F}_{1}$, and adding the result to the $\mathbf{F}_{1}$.
    The cascade process of obtaining the final output logits $\mathbf{G}_2\!\in\!\mathbb{R}^{\frac{H}{16} \times \frac{W}{16} \times 1}$ of LMC can be represented using the following equation:
\begin{small}
    \begin{align}
        \label{eq:E1}
         \mathbf{G}_{2} &= \text{Conv}(\text{Cat}(\text{U}(\mathbf{Q}_{1} \ast \mathbf{F}_{1} + \mathbf{F}_{1}), \mathbf{F}_{2} ), \theta_{2}), 
    \end{align}
\end{small}%
where the $\theta_{2}$ is the learnable parameters of the Conv. The HMC follows the same procedure to obtain the final logits $\textbf{G}_{3}\!\in\!\mathbb{R}^{\frac{H}{8} \times \frac{W}{8} \times 1}$. 
    At last, the two intermediate resolution logits $\textbf{G}_{2}$ (LMC) and $\textbf{G}_{3}$ (HMC) are fused to obtain the final Focus Map ($\mathbf{FM}\!\in\!\mathbb{R}^{\frac{H}{16} \times \frac{W}{16} \times 1}$).
\begin{small}
    \begin{align}
        \label{eq:Q_final}
        \mathbf{FM}=\text{Sigmoid}(\text{Conv}(\text{Cat}(\mathbf{G}_{2}, \mathbf{G}_{3}), \theta_{m})),
    \end{align}
\end{small}%
where the $\theta_{m}$ is the learnable parameters of the Conv.

\subsection{Separation Module}
\noindent\textbf{Intuition.} The region of FM shares similarities with the adjacent pixels, while the region of smoke can be perceived as an external element that is not compatible with the background. We leverage the contextual features surrounding FM to complement the features within the FM area, thereby aligning them with the neighboring features (i.e., smoke-free background). Consequently, by subtracting the original image, the foreground smoke can be extracted. 

\noindent\textbf{Overview.} 
In Separation module, we utilize the FM to obtain the smoke region and apply the inpainting technique (a.k.a Image completion) to fill the smoke area with smoke-free background using the surrounding scene (particularly the area around the smoke) as a reference. We then subtract the smoke-free background from the original smoke region to derive the pure smoke foreground. Furthermore, to preserve information and maintain consistency with high-level segmentation tasks, we separate the foreground and background at the \textit{feature level}.

\noindent \textbf{Detail.} As shown in Fig.~\ref{fig:FoSp_Pipeline}, In order to separate the foreground and background of smoke images at the feature level, the entire Separation module is composed of two weight-shared inpainters.

\noindent 1) In the top branch, the original image $\mathbf{I}$ and Focus Map \textbf{FM} are first input into the embedding layer $\text{EB}(\cdot)$ for feature integration and then encoded into a latent feature  $\mathbf{F}_{1}^{N}\in\mathbb{R}^{\frac{H}{32}\times \frac{W}{32} \times C_{1}}$ through the Conv block. 
\begin{small}
    \begin{align}
        \label{eq:F_N}
        \mathbf{F}_{1}^{N} = \text{Conv}(\text{EB}(\mathbf{I} ,\mathbf{FM} ),\theta_{I}), 
    \end{align}
\end{small}%
where $\theta_{I}$ represents the pre-trained Conv parameters of the inpainting network.

\noindent 2) Different from the top branch, in the bottom branch, the input is the original image $\mathbf{I}$ and a $\mathbf{Blank}$ image, without any mask.
\begin{small}
    \begin{align}
        \label{eq:F_A}
        \mathbf{F}_{1}^{A} = \text{Conv}(\text{EB}(\mathbf{I} , \mathbf{Blank}),\theta_{I}),
    \end{align}
\end{small}%
where $\mathbf{F}_{1}^{A}\in\mathbb{R}^{\frac{H}{32}\times \frac{W}{32} \times C_{1}}$.

\noindent 3) Then the latent features of the two branches are sent to three decoder blocks and three different scale inpainting-network-domain feature maps are obtained.   
\begin{small}
    \begin{align}
        \label{eq:Dec}
        \mathbf{F}_{i}^{A/N} = \text{Dec}_{i-1}(\mathbf{F} _{i-1}^{A/N}),i=2, 3, 4,
    \end{align}
\end{small}%
where $\mathbf{F}_{i}^{A/N}\in\mathbb{R}^{\frac{H}{2^{6-i}}\times \frac{W}{2^{6-i}} \times C_{i}}$.

\noindent 4) Finally, we calculate the $L^{1}$-norm of the four scale feature maps in both the top branch and the bottom branch to get the foreground features $\mathbf{F} _{i}^{F}$, forming a feature list and feeding it into the Domain Fusion module.   
\begin{small}
    \begin{align}
        \label{eq:Dec}
        \mathbf{F} _{i}^{F} = \beta \ast \Vert \mathbf{F} _{i}^{N}-\mathbf{F} _{i}^A \Vert_{1}, i=1, 2, 3 ,4
    \end{align}
\end{small}%
where $\beta$ is the gain factor and  we set $\beta=10$.

    \subsection{Domain Fusion} 

    \begin{figure}
        \begin{center}
            \includegraphics[width=\linewidth]{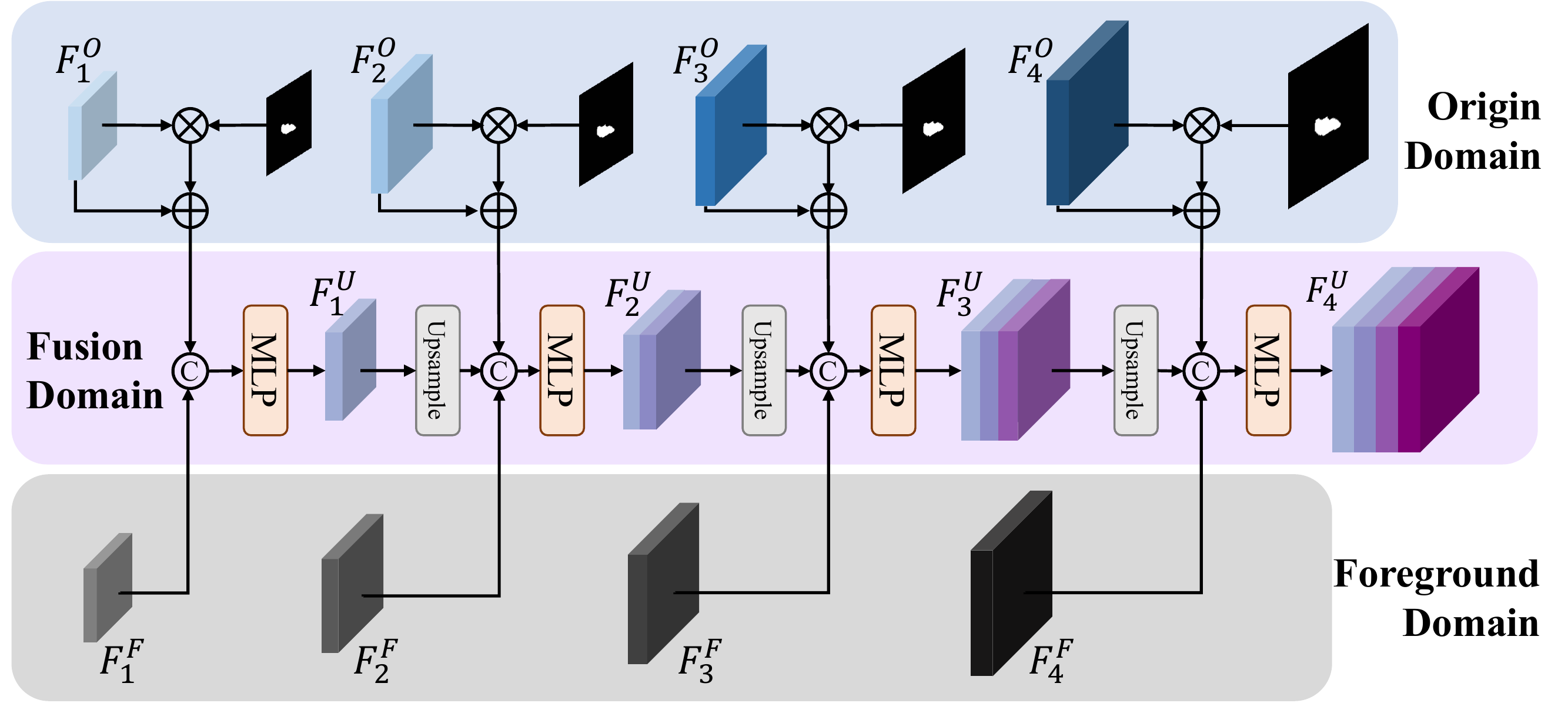}
        \end{center}
        \vspace{-2mm}
        \caption{\textbf{The structure of our Domain Fusion Module.} At each stage, the features of the two domains and the feature of previous fusion stage are combined through concatenation and then integrated using an MLP, resulting in a hierarchical merging of the features from both domains.}
        \label{fig:Fusion_Pipeline}
        \vspace{-2mm}
    \end{figure}

\noindent\textbf{Intuition.} Due to the foreground features generated by the inpainters, the origin features and foreground features actually belong to different feature domains. Simply adding features does not effectively merge the high-quality features of both, and this operation may even damage the original feature domain due to different information represented by different channels. Therefore, we design Domain Fusion module to achieve the fusion of both domain features.

\noindent \textbf{Detail.} As shown in Fig.~\ref{fig:Fusion_Pipeline}, the Domain Fusion module is divided into three domains: origin domain, foreground domain, and fusion domain. The origin domain is composed of features at different scales generated by the Focus module, and we utilize the Focus Map to enhance the features of the smoke region. The foreground domain is composed of four distinct scale foreground features generated by the Separation module. 
In each stage, the features of the two domains and the 
 feature of \textit{previous fusion stage} are concatenated and then fused with an MLP. The features of the two domains are hierarchically merged in this way.

    \subsection{Loss Function}
    The loss function consists of two parts: base loss and \textit{Focus} loss. 
    In conjunction with our proposed Focus module, we introduce a \textit{Focus} loss, which provides supervisory information to the prediction logits $\mathbf{G}_{i} (i=1, 2, 3, 4)$ of multiple resolution feature maps.
    \begin{small}
    \begin{align}
        \label{eq:loss_focus}
            \mathcal{L}_{focus} =\lambda_{f} \mathcal{L}_{BCE}(\text{FM}, \mathcal{Y} ) + \sum_{i=1}^{4} \lambda_{i}(\mathcal{L}_{BCE}(\mathbf{G}_{i},\mathcal{Y})), 
    \end{align}
\end{small}%
where $\mathcal{L}_{BCE}$ is the binary cross-entropy, $\mathcal{Y}$ is the ground truth, $\lambda_{f}$ and $\lambda_{i}$ are the weighting factors and we set $\lambda_{f}=\lambda_{i}=0.1$.

    For the basic segmentation loss, we use the standard binary cross-entropy. And the final objective function is the sum of the two loss:
\begin{small}
    \begin{align}
        \label{eq:loss}
        \mathcal{L}= \mathcal{L}_{focus} + \lambda_{b}\mathcal{L}_{BCE}(\mathbf{P} , \mathcal{Y}),
    \end{align}
\end{small}%
where $\mathbf{P}$ is the final prediction of model, $\lambda_{b}$ is the weighting factor and we set $\lambda_{b}=0.5$.

\section{Experiments}
    In this section, we first introduce our SmokeSeg dataset and training settings. Then, we compare our method with others on three datasets. Next, the ablation experiments are performed on FoSp. Finally, we further investigate some specific modules in more detail.
    \subsection{SmokeSeg Dataset}
    
    \begin{figure}[t]
    \begin{center}
        \includegraphics[width=0.88\linewidth]{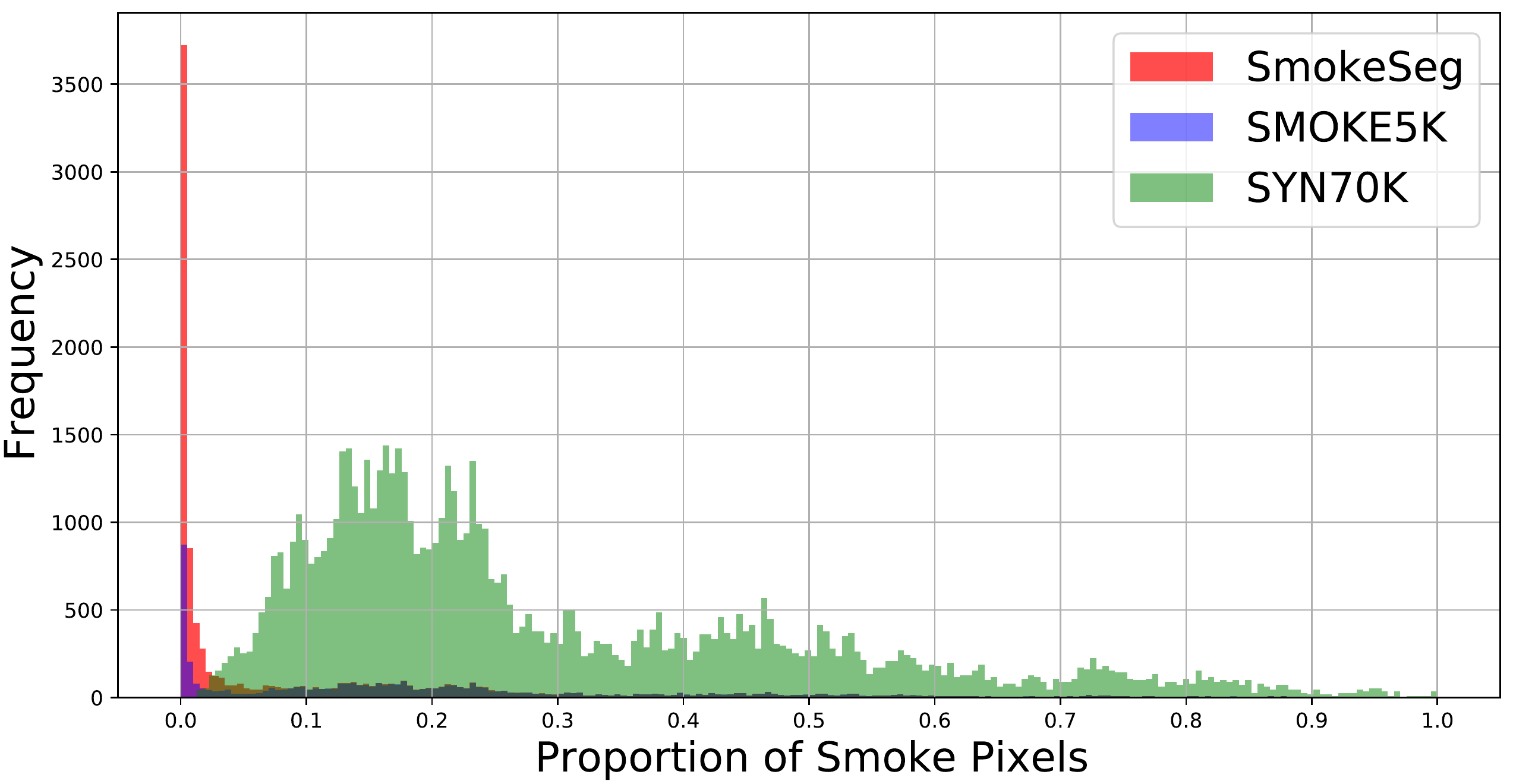}
    \end{center}
    \vspace{-2mm}
       \caption{\textbf{Distribution map of smoke pixel ratio in an image of three datasets.} The distribution of our SmokeSeg is significantly skewed to the left. This indicates that SmokeSeg contains a much higher proportion of small smoke images than other two datasets.}
    \label{fig:SmokeSeg}
    \vspace{-3mm}
    \end{figure}
    
    There are two main smoke segmentation datasets available: SYN70K~\cite{yuan2019deep} and SMOKE5K~\cite{yan2022transmission}. SYN70K comprises 70k synthetic images, while SMOKE5K is a dataset containing 1,360 real images and 4k synthetic images, with the latter being selected from SYN70K. However, neither of these datasets is tailored specifically for \textit{early smoke segmentation}.
    SYN70K is a synthetic dataset with lots of images of smoke, but the smokes in it are large and prominent, which are quite different from real smoke. 
    SMOKE5K has a limited number of real smoke images, and a low proportion of these are early smoke images, making it difficult to conduct experiments specifically targeted early smoke segmentation.

    To address these issues and further promote the development of early smoke segmentation, we contribute \textit{SmokeSeg} dataset.
    SmokeSeg consists of 6,144 real images (the raw smoke images are sourced from FigLib~\cite{dewangan2022figlib}.), which has the largest number of real images with pixel-wise annotations in any publicly available smoke segmentation dataset. 
    The number of real images in SmokeSeg is 4.5 times larger than that of SMOKE5K, which only comprises 1,360 real images. The majority of these images in SmokeSeg are early smoke, characterized by small scale and transparent appearance.
    Fig.~\ref{fig:SmokeSeg} shows the distribution of smoke pixel ratios in three datasets, with smaller ratios indicating smaller smoke. It can be seen that the SYN70K dataset hardly contains small smoke images, while SMOKE5K only has a little fraction of them. In contrast, the majority of our SmokeSeg dataset is composed of small smoke images.

    \subsection{Implementation Details}
    \noindent\textbf{Datasets.} We conduct experiments on three large smoke datasets: SYN70K~\cite{yuan2019deep}, SMOKE5K~\cite{yan2022transmission} and our SmokeSeg. To make a fair comparison, we train our models on SYN70K and SMOKE5K and test them on their respective test sets. And then we provide a new benchmark on our {SmokeSeg} dataset.

    \noindent\textbf{Evaluation Metrics.}
    Following previous smoke segmentation literature~\cite{yan2022transmission, yuan2019deep}, we evaluate our model utilizing the mMse ($\mathcal{M}$) and $mIoU$ metric on SYN70K and adopt mMse ($\mathcal{M}$) and F-measure ($F_{\beta}$) on SMOKE5K. For our \textit{SmokeSeg}, we employ the \textit{three metrics} mentioned above ($F_{\beta}$, $mIoU$, $\mathcal{M}$) to comprehensively evaluate the performance of the model.

    \noindent\textbf{Training Details.} Each image is resize to $512\times 512$. Random crop and random flip  are adopted during the training. We use the AdamW~\cite{loshchilov2017decoupled} optimizer and 
    set the learning rate to 6e-5 with $0.01$ weight decay. We train 40k iterations on SMOKE5K and SmokeSeg, and 80k iterations on SYN70K, with all batch size set to $6$. The code can be found in the supplementary materials.

    \subsection{Comparison with State-of-the-Art Methods}
   \noindent\textbf{SYN70K}~\cite{yuan2019deep}. We evaluate our model on their three synthetic test sets (DS01, DS02, and DS03). As shown in Table~\ref{tab:SYN70K_mse}, our model outperforms previous state-of-the-art method Trans-BVM~\cite{yan2022transmission} on all three test sets, with significant improvement (The $\mathcal{M} \downarrow$ have decreased by 2.81\% on DS02 test set.).

\begin{table}[!htbp]
    \centering
    \resizebox{\linewidth}{!}{
    \begin{tabular}{ccccccc}
    \toprule[1pt]
 & \multicolumn{2}{c}{DS01} & \multicolumn{2}{c}{DS02} & \multicolumn{2}{c}{DS03} \\ \midrule
Method & $\mathcal{M}\downarrow$ & $mIoU\uparrow$ & $\mathcal{M}\downarrow$ & $mIoU\uparrow$ & $\mathcal{M}\downarrow$ & $mIoU\uparrow$ \\ \midrule
SMD~\cite{wang2017video} & 0.3209 & 62.88 & 0.3379 & 61.50 & 0.3255 & 62.09 \\
TBFCN~\cite{zhang2016multi} & 0.3021 & 66.67 & 0.3196 & 65.85 & 0.3070 & 66.20 \\
LRN~\cite{islam2017label} & 0.3069 & 66.43 & 0.3078 & 67.71 & 0.3041 & 67.46 \\
HG-Net2~\cite{newell2016stacked} & 0.3186 & 64.27 & 0.3380 & 63.06 & 0.3273 & 64.18 \\
HG-Net8~\cite{newell2016stacked} & 0.3187 & 62.10 & 0.3301 & 61.89 & 0.3215 & 62.45 \\
ESPNet~\cite{mehta2018espnet} & - & 61.85 & - & 61.90 & - & 62.77 \\
LKM~\cite{peng2017large} & 0.2658 & 75.82 & 0.2799 & 74.93 & 0.2748 & 75.39 \\
RefineNet~\cite{lin2017refinenet} & 0.2486 & 77.16 & 0.2590 & 76.75 & 0.2515 & 77.52 \\
PSPNet~\cite{zhao2017pyramid} & 0.2366 & 78.71 & 0.2480 & 78.01 & 0.2430 & 78.39 \\
CCL~\cite{ding2018context} & 0.2349 & 78.87 & 0.2498 & 77.95 & 0.2429 & 78.55 \\
DFN~\cite{yu2018learning} & 0.2269 & 80.87 & 0.2411 & 79.90 & 0.2332 & 80.60 \\
DSS~\cite{yuan2019deep} & 0.2745 & 71.04 & 0.2894 & 70.01 & 0.2861 & 69.81 \\
W-Net~\cite{yuan2019wave} & 0.2688 & 73.06 & 0.2548 & 73.97 & 0.2596 & 73.36 \\
CCENet~\cite{yuan2022cubic} & - & 74.67 & - & 75.24 & - & 76.01 \\
Trans-BVM~\cite{yan2022transmission} & 0.1210 & - & 0.1362 & - & 0.1291 & - \\
\cellcolor{Gray}\textbf{FoSp (ours)} & \cellcolor{Gray}\textbf{0.1026} & \cellcolor{Gray}\textbf{83.00} & \cellcolor{Gray}\textbf{0.1081} & \cellcolor{Gray}\textbf{81.81} & \cellcolor{Gray}\textbf{0.1048} & \cellcolor{Gray}\textbf{82.80}
    \\ \bottomrule[1pt]
    \end{tabular}}
    \vspace{0.5em}
    \caption{Performance on  three test sets of SYN70K}
    \label{tab:SYN70K_mse}
    \vspace{-1mm}
\end{table}

    \noindent\textbf{SMOKE5K}~\cite{yan2022transmission}. We evaluate our model on their test set of 400 real smoke images. As demonstrated in Table~\ref{tab:SMOKE5K}, our approach surpasses previous state-of-the-art Trans-BVM~\cite{yan2022transmission} and achieves 2.5\% improvement on $F_{\beta}$.

\begin{table}[tbp]
    \centering
    \resizebox{0.50\linewidth}{!}{
    \begin{tabular}{cccccccccc}
    \toprule[1pt]
Method       & $F_{\beta}\uparrow$ & $\mathcal{M}\downarrow$  \\ \midrule
F3Net~\cite{wei2020f3net}    & 67.0     & 0.004 \\
BASNet~\cite{qin2019basnet}& 73.3     & 0.005 \\
SCRN~\cite{wu2019stacked}& 76.9     & 0.003 \\
ITSD~\cite{zhou2020interactive}& 77.4     & 0.003 \\
UCNet~\cite{zhang2020uc}& 78.7     & 0.003 \\
Trans-BVM~\cite{yan2022transmission}& 79.1     & 0.002 \\
\cellcolor{Gray}\textbf{FoSp (ours)} & \cellcolor{Gray}\textbf{81.6}   & \cellcolor{Gray}\textbf{0.002}
    \\ \bottomrule[1pt]
    \end{tabular}}
    \vspace{0.7em}
    \caption{Performance on the test set of SMOKE5K.}
    \label{tab:SMOKE5K}
    \vspace{-4mm}
\end{table}

\begin{table*}[!ht]
    \centering
    \resizebox{\linewidth}{!}{
    \begin{tabular}{ccccccccccccccc}
    \toprule[1pt]
             & \multicolumn{3}{c}{Total} & \multicolumn{3}{c}{Small} & \multicolumn{3}{c}{Medium} & \multicolumn{3}{c}{Large} &          \\ \midrule
Method       & $F_{\beta}\uparrow$   & $mIoU\uparrow$   & $\mathcal{M}\downarrow$    &$F_{\beta}\uparrow$   & $mIoU\uparrow$   & $\mathcal{M}\downarrow$    & $F_{\beta}\uparrow$   & $mIoU\uparrow$   & $\mathcal{M}\downarrow$ & $F_{\beta}\uparrow$   & $mIoU\uparrow$   & $\mathcal{M}\downarrow$   & Params (M) \\ \midrule
FCN-s~\cite{long2015fully}& 54.12  & 42.06  & 0.0105  & 33.93  & 25.00  & 0.0017  & 64.08   & 50.65  & 0.0051  & 67.24  & 52.96  & 0.0271  & 9.71     \\
DeepLabv3+-s~\cite{chen2018encoder} & 59.56  & 46.71  & 0.0095  & 44.06  & 32.77  & 0.0014  & 68.35   & 54.76  & 0.0046  & 68.37  & 54.46  & 0.0246  & 15.23    \\
PSPNet-s~\cite{zhao2017pyramid}& 57.28  & 44.76  & 0.0096  & 40.86  & 30.70  & 0.0015  & 63.77   & 50.09  & 0.0051  & 69.72  & 55.68  & 0.0245  & 13.61    \\
OCRNet-s~\cite{yuan2020object}& 62.64  & 50.11  & \textbf{0.0087}  & 50.99  & 39.47  & 0.0013  & 71.47   & 55.09  & 0.0043  & 70.45  & \textbf{57.33}  & \textbf{0.0225}  & 12.07  \\
SegFormer-s~\cite{xie2021segformer}& 60.73  & 48.19  & 0.0097  & 48.21  & 35.18  & 0.0015  & 70.22   & 56.81  & 0.0042  & 67.52  & 54.18  & 0.0256  & 6.38     \\
\cellcolor{Gray}\textbf{FoSp-s (ours)}      & \cellcolor{Gray}\textbf{64.26}  & \cellcolor{Gray}\textbf{51.44}  & \cellcolor{Gray}0.0089  & \cellcolor{Gray}\textbf{51.35}  & \cellcolor{Gray}\textbf{39.57}  & \cellcolor{Gray}\textbf{0.0012}  & \cellcolor{Gray}\textbf{72.47}   & \cellcolor{Gray}\textbf{58.95}  & \cellcolor{Gray}\textbf{0.0041}  & \cellcolor{Gray}\textbf{70.60}  & \cellcolor{Gray}{57.31}  & \cellcolor{Gray}{0.0234}  & \cellcolor{Gray}\textbf{5.79}     \\ \midrule
FCN~\cite{long2015fully}& 65.41  & 51.89  & 0.0089  & 55.30  & 41.58  & 0.0013  & 70.72   & 57.31  & 0.0043  & 71.64  & 58.22  & 0.0233  & 49.48    \\
PSPNet~\cite{zhao2017pyramid}& 65.80  & 52.42  & 0.0086  & 55.27  & 42.01  & 0.0012  & 71.44   & 58.13  & 0.0042  & 72.15  & 58.54  & 0.0224  & 48.96    \\
EncNet~\cite{zhang2018context}& 66.54  & 53.15  & 0.0088  & 58.09  & 44.86  & 0.0012  & 71.16   & 57.74  & 0.0042  & 71.54  & 57.96  & 0.0232  & \textbf{35.87}    \\
DeepLabv3+~\cite{chen2018encoder}& 67.26  & 53.85  & 0.0084  & 57.00  & 43.70  & 0.0013  & 73.34   & 59.80  & 0.0039  & 72.79  & 59.39  & 0.0219  & 43.58    \\
CCNet~\cite{huang2019ccnet}& 64.45  & 51.42  & 0.0090  & 52.94  & 40.59  & 0.0015  & 69.79   & 56.26  & 0.0046  & 72.31  & 59.01  & 0.0231  & 49.81    \\
OCRNet~\cite{yuan2020object}& 65.13  & 52.66  & 0.0085  & 52.60  & 41.04  & 0.0012  & 72.24   & 59.34  & 0.0039  & 72.25  & 59.16  & 0.0224  & 70.37    \\
SegFormer~\cite{xie2021segformer}& 67.99  & 54.98  & 0.0088  & 58.32  & 45.72  & 0.0017  & 74.34   & 61.37  & 0.0038  & 72.50  & 58.95  & 0.0235  & 44.60     \\
\cellcolor{Gray}\textbf{FoSp (ours)}        & \cellcolor{Gray}\textbf{72.05}  & \cellcolor{Gray}\textbf{59.03}  & \cellcolor{Gray}\textbf{0.0079}  & \cellcolor{Gray}\textbf{66.03}  & \cellcolor{Gray}\textbf{52.68}  & \cellcolor{Gray}\textbf{0.0010}  & \cellcolor{Gray}\textbf{75.90}   & \cellcolor{Gray}\textbf{62.99}  & \cellcolor{Gray}\textbf{0.0037}  & \cellcolor{Gray}\textbf{74.98}  & \cellcolor{Gray}\textbf{62.24}  & \cellcolor{Gray}\textbf{0.0208}  & \cellcolor{Gray}{47.52} 
    \\ \bottomrule[1pt]
    \end{tabular}}
    \vspace{0.3em}
    \caption{Performance on the test set of SmokeSeg.}
    \label{tab:SmokeSeg}
    \vspace{-1mm}
\end{table*}

    \noindent\textbf{SmokeSeg}. As there is no complete open source smoke segmentation training code available, we have chosen several strong baselines that perform well in semantic segmentation for comparison. In order to better explore the recognition abilities of various methods for smoke of different scales, especially for early smoke, we divide the test set into three parts based on the proportion of smoke pixels in an image, namely small, medium, and large, and test them separately. We use the evaluation metric on "small" to measure the \textit{early smoke segmentation capability} of our model.
    \begin{equation} \label{eq:split}{
    \begin{cases}
    \text{Small}:\delta   < 0.5\%
     \\
    \text{Medium}: 0.5\% < \delta  < 2.5\%
     \\
    \text{Large}: \delta > 2.5\%
    \end{cases}, 
    }
    \end{equation}
    \noindent where $\delta$ is the smoke pixel ratio in an image. In addition, to ensure the segmentation performance on large smoke, we also incorporated 4k synthetic smoke images into the training process.
    
    As shown in Table~\ref{tab:SmokeSeg}, the comparison is divided into two parts, with the top part of the table showing the comparison of methods with lightweight backbones, and the bottom part showing the comparison of methods with medium-sized backbones. From the comparison of the lightweight methods, our FoSp using the least number of parameters surpasses other methods with higher parameter counts in the vast majority of cases. In the comparison of the medium-sized methods, our FoSp significantly outperforms all other methods, especially in the comparison of \textit{small smoke}, where our method is 7.71\% higher than the second-best SegFormer~\cite{xie2021segformer} on $F_{\beta}$, demonstrating that our method can better recognize early smoke.

    \subsection{Ablation Study}
\begin{table*}[ht!]
    \centering
    \resizebox{\linewidth}{!}{
    \begin{tabular}{ccccccccccccc}
    \toprule[1pt]
    & \multicolumn{5}{c}{Module}                                 & \multicolumn{3}{c}{Metrics}                    & \multicolumn{3}{c}{Metrics (small)}              \\ \midrule
    Method & Baseline & $\mathcal{L}_{focus}$ & Focus & Separation & Domain Fusion & Recall  $\uparrow$  & Precision $\uparrow$    & $F_{\beta}$ $\uparrow$             & Recall  $\uparrow$       & Precision $\uparrow$     & $F_{\beta}$  $\uparrow$           \\ \midrule
(a) & \checkmark      &            &       &            &               & 71.77         & 71.34         & 67.99          & 60.82          & 63.29          & 58.32          \\
(b) & \checkmark      & \checkmark        &       &            &               & 73.53         & 71.55         & 68.90          & 65.06          & 63.06          & 60.99          \\
(c) & \checkmark      & \checkmark        & \checkmark   &            &               & \textbf{75.60} & 71.50          & 70.00             & \textbf{68.52} & 62.86          & 62.23          \\
(d) & \checkmark      & \checkmark        & \checkmark   & \checkmark        &               & 73.82         & \textbf{75.50} & 71.50           & 66.24          & \textbf{71.16} & 65.77          \\
\cellcolor{Gray}(e) & \cellcolor{Gray}\checkmark      & \cellcolor{Gray}\checkmark        & \cellcolor{Gray}\checkmark   & \cellcolor{Gray}\checkmark        & \cellcolor{Gray}\checkmark           & \cellcolor{Gray}74.60          & \cellcolor{Gray}75.20          & \cellcolor{Gray}\textbf{72.05} & \cellcolor{Gray}68.24          & \cellcolor{Gray}68.56          & \cellcolor{Gray}\textbf{66.03}
    \\ \bottomrule[1pt]
    \end{tabular}}
    \vspace{0.3em}
    \caption{Ablation study of our FoSp.}
    \label{tab:Ablation}
    \vspace{-4mm}
\end{table*}
    To further investigate the role of each module in our proposed FoSp, we conduct ablation experiments on the SmokeSeg dataset. As shown in Table~\ref{tab:Ablation}, we utilize SegFormer~\cite{xie2021segformer} as the baseline for our study. In order to investigate the influence of individual modules on the miss detection rate and precision of early smoke detection, we augment the evaluation metrics by including recall (for miss detection rate) and precision. $F_{\beta}$ is adopted as the comprehensive evaluation metric.
    \begin{figure}[t]
    \begin{center}
        \includegraphics[width=0.95\linewidth]{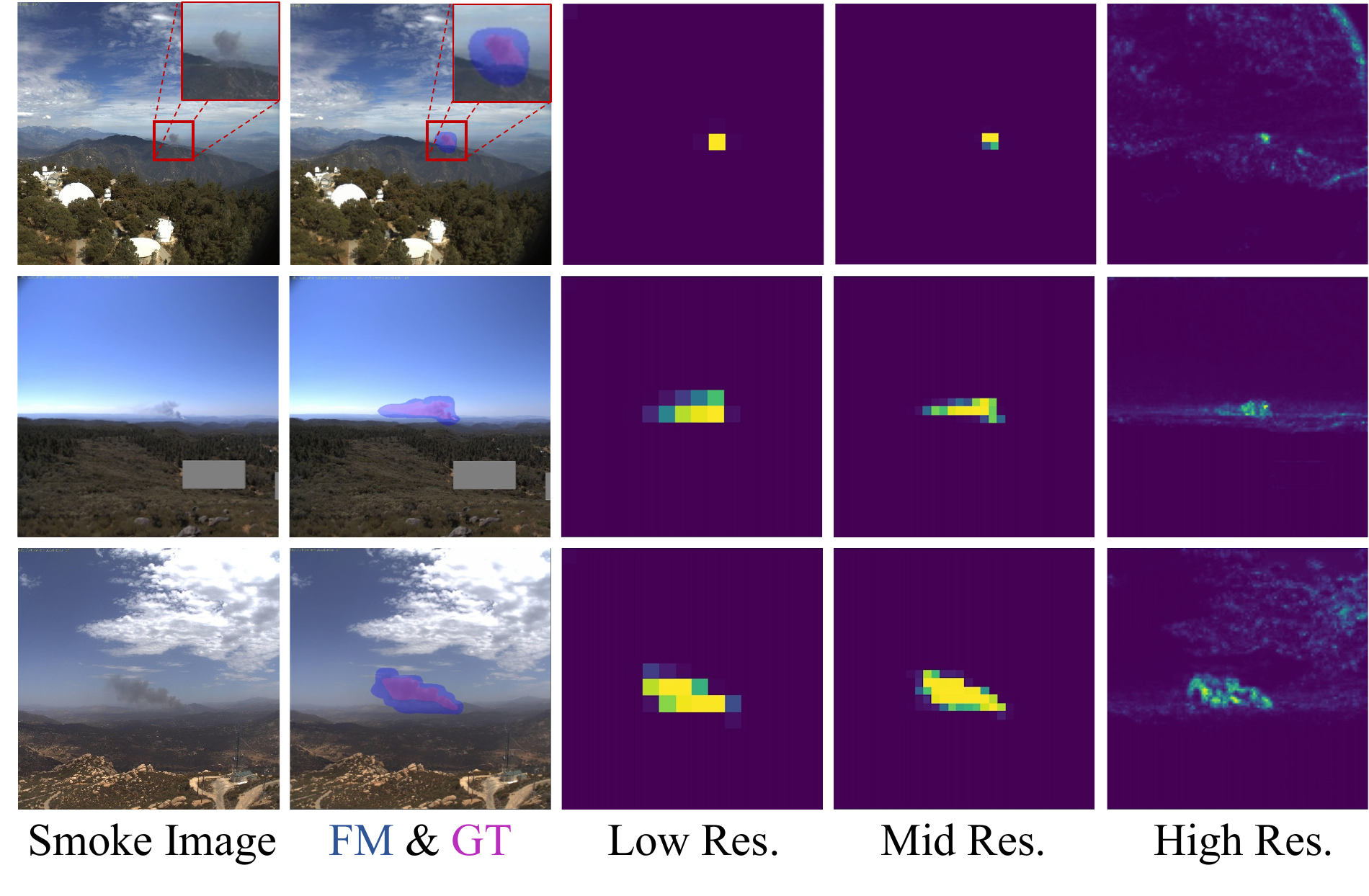}
    \end{center}
    \vspace{-4mm}
       \caption{\textbf{Performances of Focus Module.} The Focus Map (FM \& GT, The blue region represents the Focus Map, while the red denotes the ground truth.) can recognize small smoke and cover smoke of different scales. Low-resolution feature can help to cover the smoke body, while high-resolution feature provide detailed information about smoke edge.}
    \label{fig:focus_vis}
    \vspace{-4mm}
    \end{figure}
    
    \noindent\textbf{Effect of Focus Module.} Methods (b) and (c) in Table~\ref{tab:Ablation} demonstrate the role of the Focus loss and Focus module. It can be observed that after incorporating the Focus loss and Focus module, a substantial enhancement on recall is observed, particularly in the segmentation of small smoke particles, where the recall increased by 7.70\% compared to the baseline. As illustrated in Fig.~\ref{fig:focus_vis}, high-resolution features can capture fine details of the smoke edges, whereas low-resolution features can capture the overall characteristics of the smoke body. By establishing bidirectional cascade between these two types of features, a Focus map can be generated to fully cover the smoke area (GT) and suppress background interference.
    
\begin{figure*}
        \begin{center}
        \includegraphics[width=\linewidth]{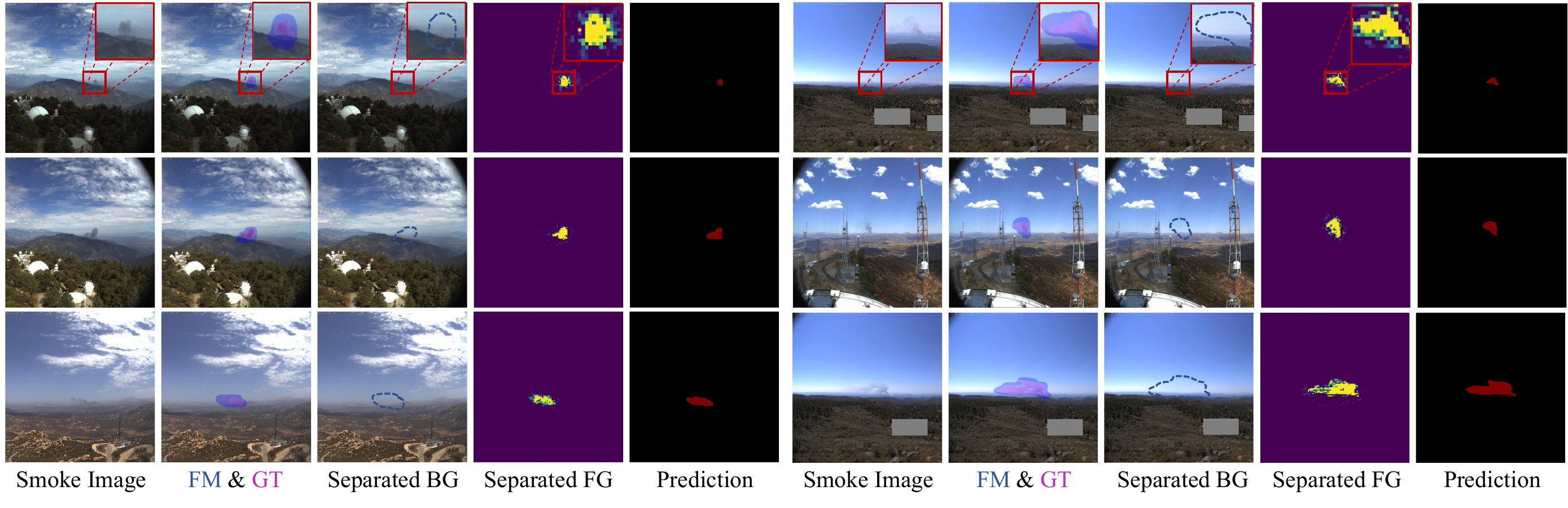}
        \end{center}
        \vspace{-2mm}
        \caption{\textbf{Performances of Separation Module.} We have visualized the Focus Map and ground truth (FM \& GT, The blue region represents the Focus Map, while the red denotes the ground truth.), the separated background image (Separated BG), the separated foreground features (Separated FG), and the prediction of FoSp. The separated background image is presented to demonstrate the effectiveness of separation which does not be used in training or testing. It can be observed that our Separation module can obtain finer foreground for both small and transparent smoke.}
        \label{fig:separation_vis}
        \vspace{-1mm}
    \end{figure*}

    \noindent\textbf{Effect of Separation Module.} Method (d) in Table~\ref{tab:Ablation} demonstrates the role of the Separation module. This method of separating the \textit{smoke foreground} from a smoke-free background results in a notable increase in precision, particularly in the segmentation of small smoke particles, where precision improved by 7.87\% compared to the baseline and by 8.30\% compared to the focus module only. The foreground feature in Fig.~\ref{fig:separation_vis}  demonstrates that the extracted smoke foreground feature closely resembles the shape of the smoke itself, particularly the fine edges of the smoke. 
    To further explore the role of the foreground features, we conduct a experiment that only the foreground features are used for prediction in the decoder head. As shown in Table~\ref{tab:FoSp_image-level}, the "Foreground only" method even outperforms the Segformer~\cite{xie2021segformer} which using the full original features on $F_{\beta}$ metric. This demonstrates the immense potential of the foreground feature in our FoSp.

    \noindent\textbf{Effect of Domain Fusion Module.} Method (e) in Table~\ref{tab:Ablation} has demonstrated the effectiveness of Domain Fusion. 
    After incorporating the Focus module, the model is inclined towards improving recall, while adding the Separation module tends to improve precision. Furthermore, after including the domain fusion module, a better balance between these two metrics is achieved, resulting in the best performance on $F_{\beta}$ metric. Compared to the baseline, the model shows a significant improvement of 4.09\% ($F_{\beta}$) on the entire test set, and an even more pronounced improvement of 7.71\% on the test set of small smoke particles.

    \subsection{Further Analysis}
\begin{table}[!htbp] 
    \centering
    \resizebox{0.75\linewidth}{!}{
    \begin{tabular}{cccccccccc}
    \toprule[1pt]
Method              & $F_{\beta}$ $\uparrow$   & $mIoU\uparrow$ &  $\mathcal{M}\downarrow$ \\ \midrule
Segformer           & 67.99 & 54.98 & 0.0088 \\
Foreground only           & 68.22 & 54.94 & 0.0087 \\
FoSp (image-level) & 71.04 & 57.81 & 0.0082 \\
\cellcolor{Gray}FoSp (feature-level)   & \cellcolor{Gray}\textbf{72.05} &\cellcolor{Gray}\textbf{59.03} &\cellcolor{Gray}\textbf{0.0079}

    \\ \bottomrule[1pt]
    \end{tabular}}
    \vspace{0.5em}
    \caption{Experiments on different separation level.}
    \label{tab:FoSp_image-level}
    \vspace{-2.0mm}
\end{table}

\begin{table}[ht!] 
    \centering
    \resizebox{0.85\linewidth}{!}{
    \begin{tabular}{cccccccccc}
    \toprule[1pt]
             & FM Threshold & FM Recall & $F_{\beta}$ $\uparrow$   & $mIoU\uparrow$ \\ \midrule
Segformer    & -            & -         & 67.99 & 54.98 \\
FoSp        & 0.9          & 42.53     & 69.79 & 55.58 \\
FoSp        & 0.8          & 53.65     & 71.75 & 58.30 \\
FoSp        & 0.7          & 62.40     & 71.52 & 58.40 \\
FoSp        & 0.6          & 69.40     & 70.30 & 57.29 \\
\cellcolor{Gray}\textbf{FoSp (ours)} & \cellcolor{Gray}\textbf{0.15} & \cellcolor{Gray}\textbf{88.74}    & \cellcolor{Gray}\textbf{72.05} & \cellcolor{Gray}\textbf{59.03}   
    \\ \bottomrule[1pt]
    \end{tabular}}
    \vspace{0.5em}
    \caption{Experiments on different Focus Map thresholds.}
    \label{tab:focus_map_recall}
    \vspace{-4mm}
\end{table}

    \noindent\textbf{Feature-level Separation and Image-level Separation.} 
    Fig.~\ref{fig:matting} has shown the performance of the image-level separation.  Although image-level separation methods can split the foreground of smoke more precisely, it is computationally expensive to restore the complete image, and downsampling is required to match the feature dimension of the original image. We have conducted experiments on both feature-level and image-level separation methods, as shown in Table~\ref{tab:FoSp_image-level}. It can be observed that the FoSp using image-level separation outperformed Segformer by 2.06\% on $F_{\beta}$ and 1.52\% on $mIoU$, which first proves the effectiveness of the FoSp concept. However, compared to the feature-level separation FoSp, the image-level FoSp results in a lower $F_{\beta}$ and $mIoU$. Therefore, we think feature-level separation is the optimal choice for our FoSp. 

\noindent\textbf{Situation When the Focus Map (FM) Can Not Cover Smoke.} 
To measure the degree of FM's coverage of smoke, we directly regard FM as the prediction and calculate its recall. In order to explore the situations where FM cannot cover the smoke, we increase the FM threshold. As presented in Table~\ref{tab:focus_map_recall}, we observe that the recall of FM decreases as the threshold value increases. 
However, even at a high threshold of 0.9, our FoSp can still reach 69.79\% on $F_{\beta}$, higher than SegFormer~\cite{xie2021segformer}. Therefore, it demonstrates that even when FM can not cover the smoke completely, FoSp still performs well. 
Meanwhile, we select an FM threshold of 0.15 in our experiments and directly use FM as the prediction for testing. The recall reaches 88.74\%, indicating that in most cases, our FM can cover the smoke.

\section{Conclusions}
    \begin{figure}[t]
    \begin{center}
        \includegraphics[width=\linewidth]{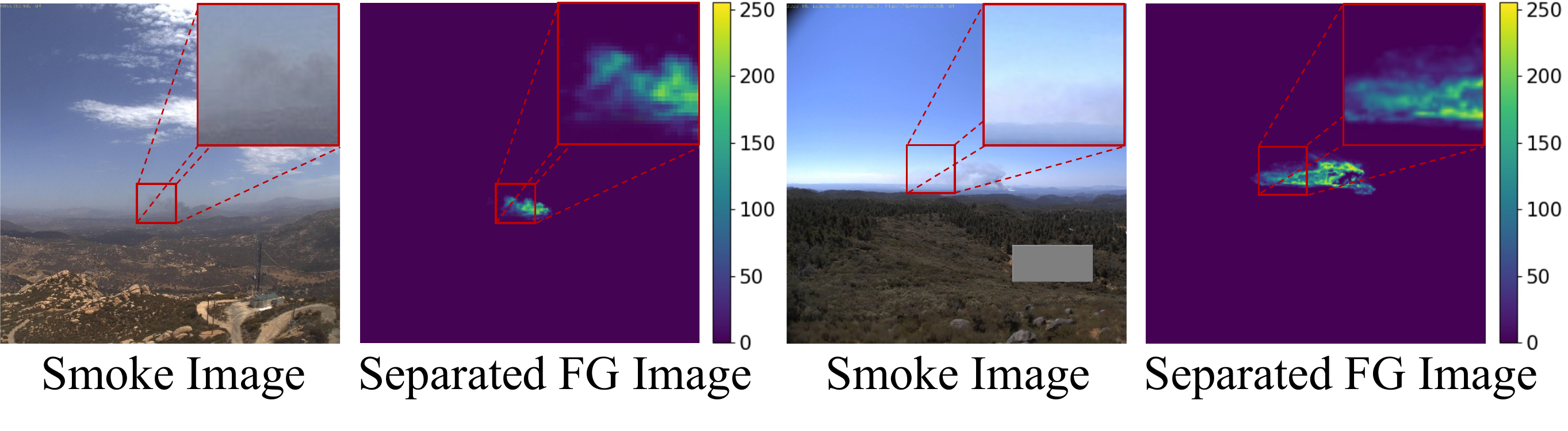}
    \end{center}
    \vspace{-4mm}
       \caption{\textbf{Performances of image-level separation.} Image-level separation (Separated FG Image) can provide clearer transparency information on smoke edge compared to the original image. The lower the value, the higher the transparency.}
    \label{fig:matting}
    \vspace{-4mm}
    \end{figure}
    In this paper, we propose a FoSp for early smoke segmentation (ESS). Concentrating on the formulation of smoke, we first determine the scope of the smoke, and then separate the pure smoke foreground from the smoke-free background, increasing the contrast between the background and foreground to obtain the complete and refined segmentation map. Furthermore, we provide a SmokeSeg dataset for ESS to promote the development of this field.
    Surprisingly, we find that sometimes using image-level separation can provide a certain degree of transparency information at the edges of smoke, as shown in Fig.~\ref{fig:matting}. This has surpassed the scope of binary segmentation and we believe that this method can be applied to more sophisticated segmentation tasks such as image matting.

{\small
\bibliographystyle{ieee_fullname}
\bibliography{ref}

\begin{thebibliography}{10}\itemsep=-1pt

\bibitem{chen2018encoder}
Liang-Chieh Chen, Yukun Zhu, George Papandreou, Florian Schroff, and Hartwig
  Adam.
\newblock Encoder-decoder with atrous separable convolution for semantic image
  segmentation.
\newblock In {\em Proceedings of the European conference on computer vision
  (ECCV)}, pages 801--818, 2018.

\bibitem{dewangan2022figlib}
Anshuman Dewangan, Yash Pande, Hans-Werner Braun, Frank Vernon, Ismael Perez,
  Ilkay Altintas, Garrison~W Cottrell, and Mai~H Nguyen.
\newblock Figlib \& smokeynet: Dataset and deep learning model for real-time
  wildland fire smoke detection.
\newblock {\em Remote Sensing}, 14(4):1007, 2022.

\bibitem{ding2018context}
Henghui Ding, Xudong Jiang, Bing Shuai, Ai~Qun Liu, and Gang Wang.
\newblock Context contrasted feature and gated multi-scale aggregation for
  scene segmentation.
\newblock In {\em Proceedings of the IEEE conference on computer vision and
  pattern recognition}, pages 2393--2402, 2018.

\bibitem{hsu2021project}
Yen-Chia Hsu, Ting-Hao~Kenneth Huang, Ting-Yao Hu, Paul Dille, Sean Prendi,
  Ryan Hoffman, Anastasia Tsuhlares, Jessica Pachuta, Randy Sargent, and Illah
  Nourbakhsh.
\newblock Project rise: recognizing industrial smoke emissions.
\newblock In {\em Proceedings of the AAAI Conference on Artificial
  Intelligence}, volume~35, pages 14813--14821, 2021.

\bibitem{huang2019ccnet}
Zilong Huang, Xinggang Wang, Lichao Huang, Chang Huang, Yunchao Wei, and Wenyu
  Liu.
\newblock Ccnet: Criss-cross attention for semantic segmentation.
\newblock In {\em Proceedings of the IEEE/CVF international conference on
  computer vision}, pages 603--612, 2019.

\bibitem{islam2017label}
Md~Amirul Islam, Shujon Naha, Mrigank Rochan, Neil Bruce, and Yang Wang.
\newblock Label refinement network for coarse-to-fine semantic segmentation.
\newblock {\em arXiv preprint arXiv:1703.00551}, 2017.

\bibitem{jia2019automatic}
Yang Jia, Hanrong Du, Haijuan Wang, Runyang Yu, Lianghui Fan, Gao Xu, and
  Qixing Zhang.
\newblock Automatic early smoke segmentation based on conditional generative
  adversarial networks.
\newblock {\em Optik}, 193:162879, 2019.

\bibitem{lin2017refinenet}
Guosheng Lin, Anton Milan, Chunhua Shen, and Ian Reid.
\newblock Refinenet: Multi-path refinement networks for high-resolution
  semantic segmentation.
\newblock In {\em Proceedings of the IEEE conference on computer vision and
  pattern recognition}, pages 1925--1934, 2017.

\bibitem{long2015fully}
Jonathan Long, Evan Shelhamer, and Trevor Darrell.
\newblock Fully convolutional networks for semantic segmentation.
\newblock In {\em Proceedings of the IEEE conference on computer vision and
  pattern recognition}, pages 3431--3440, 2015.

\bibitem{loshchilov2017decoupled}
Ilya Loshchilov and Frank Hutter.
\newblock Decoupled weight decay regularization.
\newblock {\em arXiv preprint arXiv:1711.05101}, 2017.

\bibitem{mahmoud2019forest}
Mubarak Adam~Ishag Mahmoud and Honge Ren.
\newblock Forest fire detection and identification using image processing and
  svm.
\newblock {\em Journal of Information Processing Systems}, 15(1):159--168,
  2019.

\bibitem{mehta2018espnet}
Sachin Mehta, Mohammad Rastegari, Anat Caspi, Linda Shapiro, and Hannaneh
  Hajishirzi.
\newblock Espnet: Efficient spatial pyramid of dilated convolutions for
  semantic segmentation.
\newblock In {\em Proceedings of the european conference on computer vision
  (ECCV)}, pages 552--568, 2018.

\bibitem{muhammad2018early}
Khan Muhammad, Jamil Ahmad, and Sung~Wook Baik.
\newblock Early fire detection using convolutional neural networks during
  surveillance for effective disaster management.
\newblock {\em Neurocomputing}, 288:30--42, 2018.

\bibitem{najibi2019autofocus}
Mahyar Najibi, Bharat Singh, and Larry~S Davis.
\newblock Autofocus: Efficient multi-scale inference.
\newblock In {\em Proceedings of the IEEE/CVF international conference on
  computer vision}, pages 9745--9755, 2019.

\bibitem{newell2016stacked}
Alejandro Newell, Kaiyu Yang, and Jia Deng.
\newblock Stacked hourglass networks for human pose estimation.
\newblock In {\em Computer Vision--ECCV 2016: 14th European Conference,
  Amsterdam, The Netherlands, October 11-14, 2016, Proceedings, Part VIII 14},
  pages 483--499. Springer, 2016.

\bibitem{peng2017large}
Chao Peng, Xiangyu Zhang, Gang Yu, Guiming Luo, and Jian Sun.
\newblock Large kernel matters--improve semantic segmentation by global
  convolutional network.
\newblock In {\em Proceedings of the IEEE conference on computer vision and
  pattern recognition}, pages 4353--4361, 2017.

\bibitem{qin2019basnet}
Xuebin Qin, Zichen Zhang, Chenyang Huang, Chao Gao, Masood Dehghan, and Martin
  Jagersand.
\newblock Basnet: Boundary-aware salient object detection.
\newblock In {\em Proceedings of the IEEE/CVF conference on computer vision and
  pattern recognition}, pages 7479--7489, 2019.

\bibitem{robinson1979smoke}
Donald~A Robinson.
\newblock Smoke detection: Critical element of a university residential fire
  safety program.
\newblock {\em Journal of the American College Health Association},
  27(5):265--66, 1979.

\bibitem{strudel2021segmenter}
Robin Strudel, Ricardo Garcia, Ivan Laptev, and Cordelia Schmid.
\newblock Segmenter: Transformer for semantic segmentation.
\newblock In {\em Proceedings of the IEEE/CVF International Conference on
  Computer Vision}, pages 7262--7272, 2021.

\bibitem{uzkent2020efficient}
Burak Uzkent, Christopher Yeh, and Stefano Ermon.
\newblock Efficient object detection in large images using deep reinforcement
  learning.
\newblock In {\em Proceedings of the IEEE/CVF winter conference on applications
  of computer vision}, pages 1824--1833, 2020.

\bibitem{wang2014early}
Shidong Wang, Yaping He, Ju~Jia Zou, Dechuang Zhou, and Jian Wang.
\newblock Early smoke detection in video using swaying and diffusion feature.
\newblock {\em Journal of Intelligent \& Fuzzy Systems}, 26(1):267--275, 2014.

\bibitem{wang2017video}
Wenguan Wang, Jianbing Shen, and Ling Shao.
\newblock Video salient object detection via fully convolutional networks.
\newblock {\em IEEE Transactions on Image Processing}, 27(1):38--49, 2017.

\bibitem{wang2020glance}
Yulin Wang, Kangchen Lv, Rui Huang, Shiji Song, Le Yang, and Gao Huang.
\newblock Glance and focus: a dynamic approach to reducing spatial redundancy
  in image classification.
\newblock {\em Advances in Neural Information Processing Systems},
  33:2432--2444, 2020.

\bibitem{wei2020f3net}
Jun Wei, Shuhui Wang, and Qingming Huang.
\newblock F$^3$net: fusion, feedback and focus for salient object detection.
\newblock In {\em Proceedings of the AAAI conference on artificial
  intelligence}, volume~34, pages 12321--12328, 2020.

\bibitem{wu2019stacked}
Zhe Wu, Li Su, and Qingming Huang.
\newblock Stacked cross refinement network for edge-aware salient object
  detection.
\newblock In {\em Proceedings of the IEEE/CVF international conference on
  computer vision}, pages 7264--7273, 2019.

\bibitem{xie2021segformer}
Enze Xie, Wenhai Wang, Zhiding Yu, Anima Anandkumar, Jose~M Alvarez, and Ping
  Luo.
\newblock Segformer: Simple and efficient design for semantic segmentation with
  transformers.
\newblock {\em Advances in Neural Information Processing Systems},
  34:12077--12090, 2021.

\bibitem{xie2020spatially}
Zhenda Xie, Zheng Zhang, Xizhou Zhu, Gao Huang, and Stephen Lin.
\newblock Spatially adaptive inference with stochastic feature sampling and
  interpolation.
\newblock In {\em European conference on computer vision}, pages 531--548.
  Springer, 2020.

\bibitem{xing2015smoke}
Deng Xing, Yu Zhongming, Wang Lin, and Li Jinlan.
\newblock Smoke image segmentation based on color model.
\newblock {\em Journal on Innovation and Sustainability RISUS}, 6(2):130--138,
  2015.

\bibitem{yan2022transmission}
Siyuan Yan, Jing Zhang, and Nick Barnes.
\newblock Transmission-guided bayesian generative model for smoke segmentation.
\newblock 2022.

\bibitem{yang2022querydet}
Chenhongyi Yang, Zehao Huang, and Naiyan Wang.
\newblock Querydet: Cascaded sparse query for accelerating high-resolution
  small object detection.
\newblock In {\em Proceedings of the IEEE/CVF Conference on Computer Vision and
  Pattern Recognition}, pages 13668--13677, 2022.

\bibitem{yu2018learning}
Changqian Yu, Jingbo Wang, Chao Peng, Changxin Gao, Gang Yu, and Nong Sang.
\newblock Learning a discriminative feature network for semantic segmentation.
\newblock In {\em Proceedings of the IEEE conference on computer vision and
  pattern recognition}, pages 1857--1866, 2018.

\bibitem{yuan2019learning}
Chi Yuan, Zhixiang Liu, and Youmin Zhang.
\newblock Learning-based smoke detection for unmanned aerial vehicles applied
  to forest fire surveillance.
\newblock {\em Journal of Intelligent \& Robotic Systems}, 93(1):337--349,
  2019.

\bibitem{yuan2022cubic}
Feiniu Yuan, Zeshu Dong, Lin Zhang, Xue Xia, and Jinting Shi.
\newblock Cubic-cross convolutional attention and count prior embedding for
  smoke segmentation.
\newblock {\em Pattern Recognition}, 131:108902, 2022.

\bibitem{yuan2019wave}
Feiniu Yuan, Lin Zhang, Xue Xia, Qinghua Huang, and Xuelong Li.
\newblock A wave-shaped deep neural network for smoke density estimation.
\newblock {\em IEEE transactions on image processing}, 29:2301--2313, 2019.

\bibitem{yuan2019deep}
Feiniu Yuan, Lin Zhang, Xue Xia, Boyang Wan, Qinghua Huang, and Xuelong Li.
\newblock Deep smoke segmentation.
\newblock {\em Neurocomputing}, 357:248--260, 2019.

\bibitem{yuan2020object}
Yuhui Yuan, Xilin Chen, and Jingdong Wang.
\newblock Object-contextual representations for semantic segmentation.
\newblock In {\em Computer Vision--ECCV 2020: 16th European Conference,
  Glasgow, UK, August 23--28, 2020, Proceedings, Part VI 16}, pages 173--190.
  Springer, 2020.

\bibitem{zhang2018context}
Hang Zhang, Kristin Dana, Jianping Shi, Zhongyue Zhang, Xiaogang Wang, Ambrish
  Tyagi, and Amit Agrawal.
\newblock Context encoding for semantic segmentation.
\newblock In {\em Proceedings of the IEEE conference on Computer Vision and
  Pattern Recognition}, pages 7151--7160, 2018.

\bibitem{zhang2020uc}
Jing Zhang, Deng-Ping Fan, Yuchao Dai, Saeed Anwar, Fatemeh~Sadat Saleh, Tong
  Zhang, and Nick Barnes.
\newblock Uc-net: Uncertainty inspired rgb-d saliency detection via conditional
  variational autoencoders.
\newblock In {\em Proceedings of the IEEE/CVF conference on computer vision and
  pattern recognition}, pages 8582--8591, 2020.

\bibitem{zhang2016multi}
Zheng Zhang, Chengquan Zhang, Wei Shen, Cong Yao, Wenyu Liu, and Xiang Bai.
\newblock Multi-oriented text detection with fully convolutional networks.
\newblock In {\em Proceedings of the IEEE conference on computer vision and
  pattern recognition}, pages 4159--4167, 2016.

\bibitem{zhao2017pyramid}
Hengshuang Zhao, Jianping Shi, Xiaojuan Qi, Xiaogang Wang, and Jiaya Jia.
\newblock Pyramid scene parsing network.
\newblock In {\em Proceedings of the IEEE conference on computer vision and
  pattern recognition}, pages 2881--2890, 2017.

\bibitem{zhou2020interactive}
Huajun Zhou, Xiaohua Xie, Jian-Huang Lai, Zixuan Chen, and Lingxiao Yang.
\newblock Interactive two-stream decoder for accurate and fast saliency
  detection.
\newblock In {\em Proceedings of the IEEE/CVF conference on computer vision and
  pattern recognition}, pages 9141--9150, 2020.

\end{thebibliography}
}

\end{document}